\documentclass[11pt]{article}

\usepackage[final]{acl}

\usepackage{xspace}
\newcommand{\libname}{lrnnx\xspace}

\usepackage{times}
\usepackage{pifont}
\usepackage{balance}
\usepackage{amsmath}
\usepackage{latexsym}
\usepackage{booktabs}
\usepackage{cleveref}
\usepackage{makecell}
\usepackage{subcaption}

\usepackage{xcolor}
\usepackage{listings}

\definecolor{codegreen}{rgb}{0,0.6,0}
\definecolor{codegray}{rgb}{0.5,0.5,0.5}
\definecolor{codepurple}{rgb}{0.58,0,0.82}
\definecolor{backcolour}{rgb}{0.95,0.95,0.92}

\lstdefinestyle{mystyle}{
    backgroundcolor=\color{backcolour},   
    commentstyle=\color{codegreen},
    keywordstyle=\color{magenta},
    numberstyle=\tiny\color{codegray},
    stringstyle=\color{codepurple},
    basicstyle=\ttfamily\footnotesize,
    breakatwhitespace=false,         
    breaklines=true,                 
    captionpos=b,                    
    keepspaces=true,                 
    numbers=left,                    
    numbersep=5pt,  
    xleftmargin=15pt,
    showspaces=false,                
    showstringspaces=false,
    showtabs=false,                  
    tabsize=2
}

\newcommand*\samethanks[1][\value{footnote}]{\footnotemark[#1]}

\usepackage[T1]{fontenc}

\usepackage[utf8]{inputenc}

\usepackage{microtype}

\usepackage{inconsolata}

\usepackage{graphicx}

\usepackage{float}

\usepackage{multicol}
\usepackage{caption}
\usepackage[toc,page]{appendix}

\title{\texttt{\libname}: A library for Linear RNNs}

\author{
\\
 Karan Bania\thanks{Equal contribution.}\textsuperscript{1},
 Soham Kalburgi\samethanks\textsuperscript{2},
 Manit Tanwar\samethanks\textsuperscript{2},
 Dhruthi Kiran\samethanks\textsuperscript{2},
\\
 Aditya Nagarsekar\samethanks\textsuperscript{2},
 Harshvardhan Mestha\samethanks\textsuperscript{2},
 Naman Chibber\samethanks\textsuperscript{2},
 Anish Sathyanarayanan\samethanks\textsuperscript{2},
\\
 Aarush Rathore\samethanks\textsuperscript{2},
 Raj Deshmukh\samethanks\textsuperscript{2},
 Pratham Chheda\samethanks\textsuperscript{2}
\\
\\
 \textsuperscript{1}Carnegie Mellon University,
 \textsuperscript{2}BITS Pilani, K. K. Birla Goa Campus
\\
\\
 \url{https://github.com/SforAiDl/lrnnx}
}

\begin{document}
\maketitle


\begin{abstract}
Linear recurrent neural networks (LRNNs) provide a structured approach to sequence modeling that bridges classical linear dynamical systems and modern deep learning, offering both expressive power and theoretical guarantees on stability and trainability.
In recent years, multiple LRNN-based architectures have been proposed, each introducing distinct parameterizations, discretization schemes, and implementation constraints.
However, existing implementations are fragmented across different software frameworks, often rely on framework-specific optimizations, and in some cases require custom CUDA kernels or lack publicly available code altogether. As a result, using, comparing, or extending LRNNs requires substantial implementation effort.
To address this, we introduce \texttt{\libname}, a unified software library that implements several modern LRNN architectures under a common interface.
The library exposes multiple levels of control, allowing users to work directly with core components or higher-level model abstractions.
\texttt{\libname} aims to improve accessibility, reproducibility, and extensibility of LRNN research and applications.
We make our code available under a permissive MIT license.

\end{abstract}


\section{Introduction}

\begin{table*}[htbp]
    \centering
    \begin{tabular}{lcccc}
        \toprule
        Layer & SISO & LTI & \makecell{Public \\ Implementation} & Framework \\
        \midrule
        S4~\citep{gu2022efficiently} & \ding{51} & \ding{51} & \ding{51} & PyTorch \\
        S5~\citep{smith2023simplified} & \ding{55} & \ding{51} & \ding{51} & JAX \\
        LRU~\citep{orvieto2023resurrectingrecurrentneuralnetworks} & \ding{55} & \ding{51} & \ding{55} & N/A \\
        Event-SSM~\citep{Schoene2024} & \ding{55} & \ding{51} & \ding{51} & JAX \\
        S6~\citep{gu2024mamba} & \ding{51} & \ding{55} & \ding{51} & PyTorch \\
        STREAM~\citep{schöne2024streamuniversalstatespacemodel} & \ding{51} & \ding{55} & \ding{51} & PyTorch \\
        RG-LRU~\citep{de2024griffinmixinggatedlinear} & \ding{55} & \ding{55} & \ding{55} & N/A \\
        S7~\citep{soydan2024s7selectivesimplifiedstate} & \ding{55} & \ding{55} & \ding{55} & N/A \\
        Centaurus~\citep{pei2025let} & \ding{55} & \ding{55} & \ding{51} & PyTorch \\
        \bottomrule
    \end{tabular}
    \caption{\centering
    An overview of contemporary SSM architectures and their existing implementations (SISO: Single-Input Single-Output, LTI: Linear Time Invariant).}
    \label{tab:implementations}
\end{table*}


\subsection{Context and Motivation}

Recurrent neural networks (RNNs) are a classical approach to sequence modeling, which model context explicitly with a latent state. A conventional (non-linear) RNN can be described by \cref{eq:base_non_linear_rnn}:
\begin{equation}
\label{eq:base_non_linear_rnn}
    \begin{split}
        x_k &= \alpha( W_{xx} x_{k-1} + W_{xu} u_k ), \\
        y_k &= \beta( W_{yx} x_k ),
    \end{split}
\end{equation}
where $\alpha$ and $\beta$ are non-linear activation functions.
These non-linearities are largely responsible for the expressive power of RNNs, including results on Turing completeness~\citep{turing_complete}.
However, non-linear RNNs suffer from two well-known limitations: (i) the vanishing and exploding gradient problem~\citep{lstm}, which hinders both training stability and the learning of long-range dependencies, and (ii) the inherently sequential nature of training, which limits effective utilization of modern parallel hardware.

Despite these drawbacks, RNNs possess a highly desirable property: $\mathcal{O}(1)$ time complexity for inference.
Transformers~\citep{Vaswani+2017}, which have become the dominant paradigm for sequence modeling, address both gradient instability and sequential training.
However, they do so by abandoning the notion of an explicit latent state, resulting in $\mathcal{O}(n)$ time complexity for inference due to global attention, where $n$ denotes the sequence length.

Linear recurrent neural networks (LRNNs) revisit the recurrent paradigm by restricting the state update to linear dynamics while carefully controlling stability through parameterization and discretization.
This line of work has produced a family of models that combine efficient parallel training with $\mathcal{O}(1)$ inference-time complexity, while setting new records on long-range sequence modeling benchmarks.
Moreover, LRNNs possess an inductive bias for signal data, enabling efficient end-to-end modeling of high-frequency modalities such as audio and sensor data streams.


\subsection{Implementation Challenges}

While the theoretical foundations and empirical performance of LRNNs have matured over time, their practical use remains hindered by the current fragmented implementation landscape.
As illustrated in Table~\ref{tab:implementations}, existing LRNN architectures differ not only in modeling assumptions but also in software availability and framework choice.
For example, comparing two conceptually similar models may require switching between PyTorch and JAX, adapting data pipelines, and re-implementing training utilities, while reproducing reported runtimes may further depend on custom CUDA kernels or unpublished low-level optimizations.
In several cases, no public implementation is available at all, forcing researchers to re-implement entire models from scratch.
This makes it difficult to reproduce results, benchmark models under consistent conditions, or integrate LRNNs into downstream applications.
As a consequence, using LRNNs in practice or experimenting with them beyond a single architecture requires substantial engineering overhead.

\subsection{The \texttt{\libname} Library}

We address these challenges by introducing \texttt{\libname}, a unified library designed to make working with LRNNs comparable to working with standard neural network layers.
The library provides consistent implementations of multiple LRNN architectures within a single unified framework, and abstracts away model-specific engineering details.
As a result, switching between different LRNN formulations - such as changing the state-space parameterization or discretization scheme - amounts to instantiating a different class of the library, without needing to modify the surrounding training or evaluation code.
\texttt{\libname} exposes both low-level building blocks (core recurrences) and higher-level modules (with activations and skip connections), supporting fine-grained research and experimentation as well as drop-in use in existing pipelines for direct application.

Our contributions in this work include the development of \texttt{\libname}, a unified framework that standardizes fragmented LRNN architectures into a single interface supported by high-performance custom CUDA kernels, thereby bridging the gap between research and deployment while significantly reducing the engineering overhead for cross-model benchmarking.


\section{Related Work}
\label{sec:related_work}

Since the introduction of GPT-3~\cite{gpt3}, a large body of research has focused on optimizing Transformer architectures and expanding their applications to diverse domains.

\subsection{Speeding up Transformers}
Efforts to mitigate the quadratic complexity of the Transformer's self-attention mechanism have yielded several approaches.
LongFormer~\citep{beltagy2020longformer} replaces full attention with a combination of sliding window and global attention patterns.
A broader class of \textit{sub-quadratic methods} uses techniques like low-rank projections~\citep{wang2020linformer} or locality-sensitive hashing~\citep{kitaev2020reformer} to approximate attention more efficiently.
A few hardware-aware techniques have also emerged.
FlashAttention~\citep{dao2022flashattention} reduces memory I/O without any approximations and vLLM~\citep{kwon2023efficient} introduces paged attention for efficient memory management.
Recently, there has also been some work on pseudo distillation techniques like Matryoshka Embeddings~\citep{kusupati2022matryoshka} and Speculative Decoding~\citep{leviathan2023fast}.
Most of these methods are transferrable to LRNNs.

\begin{figure*}[t]
    \centering
    \includegraphics[width=\textwidth]{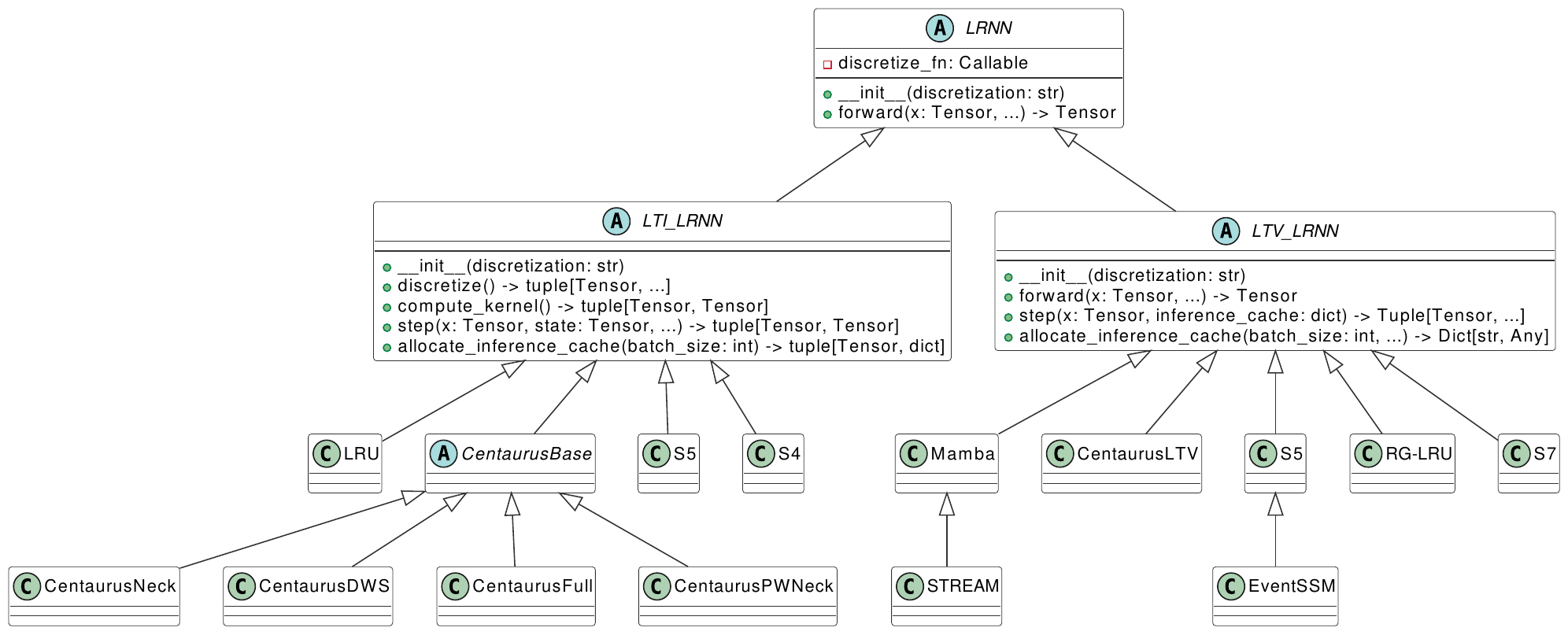} 
    \caption{Class diagram describing \texttt{\libname}.}
    \label{fig:class_diagram}
\end{figure*}

\subsection{Linear RNNs}
\label{sec:linear_rnns}
The central equation for LRNNs is described in \cref{eq:linear_rnn}.

\begin{equation}
\label{eq:linear_rnn}
    \begin{alignedat}{2}
        x_k &= A(k) x_{k-1} &&+ B(k) u_k \\
        y_k &= C(k) x_k     &&+ D(k) u_k
    \end{alignedat}
\end{equation}

Layer variants differ in how they parameterize the learnable matrices $A, B$ and $C$.
These layers can be broadly divided into two types: Linear Time Invariant (LTI) and Linear Time Varying (LTV).


\subsubsection{LTI Layers}
These layers maintain time-invariant matrices, i.e., $A(k) = A,\, \forall k$ (likewise for $B$ and $C$). 
S4~\cite{gu2022efficiently} developed much of the theory required to train and compute this recurrence efficiently.
These layers rely on the single-input single-output (SISO) framework, and use independent layers for each hidden dimension in the input.
The S5~\cite{smith2023simplified} layer extends S4 to train a multi-input multi-output (MIMO) model. 
LRU~\cite{orvieto2023resurrectingrecurrentneuralnetworks} re-formulates the problem from a deep learning perspective and develops methods to train a LRNN without the signal processing theory.
The network is similar to S5 but makes no assumptions about the input signal $u_k$.


\subsubsection{LTV Layers}
These layers have time-varying matrices, and most have a direct LTI counterpart.
S6~\cite{gu2024mamba} is a time varying variant of S4 which makes it well suited for discrete modalities like text.
S7~\cite{soydan2024s7selectivesimplifiedstate} is a time-varying variant of S5, and the RG-LRU~\cite{de2024griffinmixinggatedlinear} is a time-varying alternative to LRU.
STREAM~\citep{schöne2024streamuniversalstatespacemodel} introduces a time-varying SISO state-space model that selectively updates state components to capture varying temporal frequencies in long sequences.

Finally, Centaurus~\cite{pei2025let} is in-between SISO and MIMO models.


\subsection{Applications}
\label{sec:applications}
Overall, these layers are a rich set of architectures which have been applied to several sequential and non-sequential domains from Audio (Text-to-speech~\citep{pmlr-v162-goel22a}, ASR~\citep{pei2025let},  Enhancement~\citep{pei2025let,pei2025atennuateoptimizedrealtimespeech}), RNA  modeling~\citep{ramesh2025lyraefficientexpressivesubquadratic}, Vision~\citep{liu2024vmamba}, Event-streams~\citep{Schoene2024} and even Point-clouds~\citep{mamba3d}.
Furthermore, they have set new benchmarks on synthetic tasks in the long-range-arena (LRA)~\citep{tay2021long}.
Typically, transformers are hard to train for very long sequences ($\ge 2^{10}$), which is where these layers prove extremely useful.


\section{Library Design}

This section provides a high-level overview of \texttt{\libname}, describing its software architecture and core design principles.

Each layer in \texttt{\libname} follows a consistent interface derived from \cref{eq:linear_rnn}.
Model-specific details are abstracted behind a unified API for instantiation, training, and inference across all LRNN architectures.
A summary of supported layer architectures is provided in Table~\ref{tab:implementations}.

We adopt a three-tier inheritance hierarchy. At the base, the \texttt{LRNN} class defines the forward interface and selects the discretization method.
Layers are organized into LTI and LTV submodules corresponding to the variants described in \cref{sec:linear_rnns}. 
LTI layers extend the \texttt{LTI\_LRNN} class.
For these layers, we implement optimal einsum contractions~\citep{pei2025atennuateoptimizedrealtimespeech}, which lead to efficiency gains.
LTV layers extend the \texttt{LTV\_LRNN} class.
Each subclass defines its own parameterization of the matrices $(A, B, C)$ from \cref{eq:linear_rnn}, while preserving a shared programming interface.
The broad layout of the library is as indicated in Figure~\ref{fig:class_diagram}.

Layer definition is decoupled from discretization.
Supported schemes include ZOH, bilinear, dirac, and asynchronous (event-driven) discretization.
Some models restrict supported methods (e.g., Centaurus uses only ZOH), and the design allows easy integration of custom schemes.

Layers follow a uniform constructor signature. For example, an S5 layer can be instantiated as:
\begin{lstlisting}[language=Python]
layer = S5(
       d_model=512, 
       d_state=64,
       discretization="zoh",
       **kwargs
)
\end{lstlisting}

For efficient autoregressive generation, all layers implement a \texttt{step} method. 

For time-varying layers, \texttt{\libname} provides custom CUDA kernels, derived from the selective scan implementation in Mamba~\citep{gu2024mamba}.
These kernels integrate multiple discretization methods (ZOH, bilinear, dirac) and support asynchronous inputs within a fused scan and output projection, preserving memory efficiency while enabling flexible architectural choices.
This is a benefit over some JAX implementations, which, while easy to implement, suffer from memory bottlenecks due to materialization of the hidden state.

To ensure correctness, we validate numerical equivalence between parallel, recurrent, and step-wise execution modes for every layer with an extensive and robust test suite, across sequence lengths, batch sizes, model dimensions, initializations, and discretizations.
We further verify gradient consistency between custom CUDA kernels and reference PyTorch implementations.


\subsection{Tutorials \& Architectures}
For end-to-end applications, the library provides components and tutorials for tasks such as language modeling, classification, and autoencoders.
For example, \texttt{LRNNLMHeadModel} wraps an LRNN backbone with embeddings, stacked residual blocks, and a language modeling head:

\begin{lstlisting}[language=Python]
lm = LRNNLMHeadModel(
    d_model=768, d_state=16, n_layer=12,
    vocab_size=50257, 
    mixer_types=["S5", "S7", "attn", ...],
    mixer_kwargs={"S5": {..}, "S7": {..}, "attn": {..}, ...}, 
    d_intermediate=2048,
)
\end{lstlisting}

This design mirrors the head abstractions used in modern deep learning frameworks like Transformers~\citep{hftransformers}, enabling flexible adaptation to downstream tasks.
The \texttt{mixer\_types} argument allows mixing different LRNN backends and attention layers~\citep{de2024griffinmixinggatedlinear}, while blocks, normalization, and MLP components remain fully configurable.
All layers integrate with standard PyTorch workflows, including checkpointing, gradient checkpointing, mixed-precision training, and fused operations.


\subsection{Inference support}
JAX provides native support for such models with the \texttt{jax.lax.scan} operation which can remove CPU overheads entirely from the generation process.
Analogues of this functionality do not exist in PyTorch, and a simple for-loop would give up all the benefits of fast inference.
To mitigate this, similar to~\citet{gu2024mamba}, we provide specialized inference capabilities using CUDA Graphs, to avoid CPU synchronization after each step.
Our implementation is competitive at large sequence lengths and only adds a few ms at small ones.


\section{Experiments}


\begin{figure*}[t]
    \centering
    \includegraphics[width=\textwidth]{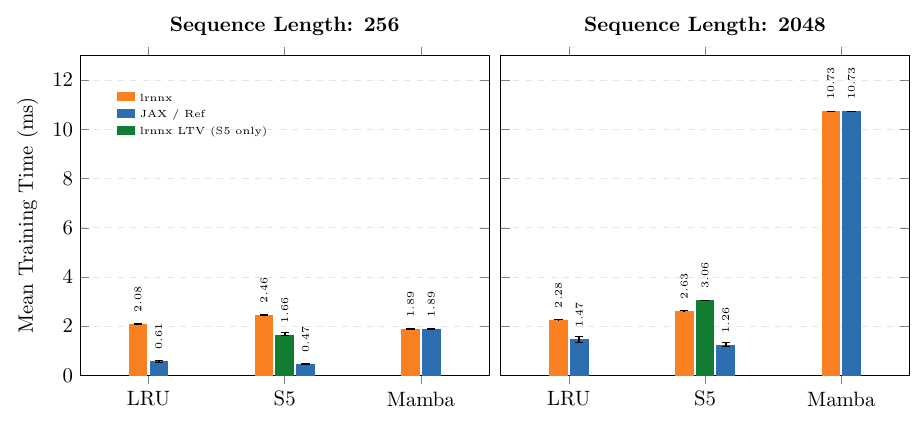} 
    \caption{Training Time Comparison}
    \label{fig:benchmarks_main}
\end{figure*}

\subsection{Setup}
We run all of our GPU benchmarks on an NVIDIA A100 40GB GPU, using Python 3.12 and CUDA 12.9.

\subsection{Benchmarks}
We provide a performance analysis of our \texttt{\libname} implementations by comparing them to their original or alternative counterparts, on random tensors.
We have evaluated our LRU implementation (PyTorch) against a popular public repository~\citep{zucchet2023minimalLRU} (JAX).
Our S5 implementation was compared against the original release~\citep{smith2023simplified}, and similarly the Mamba implementation is evaluated relative to the official repository~\citep{gu2024mamba}.

We report average execution time (ms) for both training (forward plus backward pass) and autoregressive inference across models while varying batch size, sequence length, and model dimension for LRU, S5, and Mamba.
For each configuration we run 10 warm‑up passes, then time 90 forward passes; this is repeated for 5 experiments, and we report the mean and standard deviation across those 5 experiment means.
We mirror the same sweep settings across all three models (batch sizes, sequence lengths, and model dimensions), and all plots use log scaling where specified.
Wherever required, we set the state dimension to 16.
Overall, our implementations are competitive to the public baselines -- \Cref{fig:benchmarks_main}.
All benchmark results can be found in Appendix~\ref{app:benchmarks}.


\section{Conclusion}

In this work, we introduce \texttt{\libname}, a unified library consolidating SOTA linear RNN architectures into a single interface.
By providing $O(1)$ inference complexity and strong inductive biases for signal-like data, the library facilitates efficient long-sequence modeling across diverse domains, including audio, vision, and event-streams (\cref{sec:applications}).
We expect \texttt{\libname} to empower the community with a scalable, easily extensible, and accessible alternative where Transformer-based methods encounter limitations.


\section*{Limitations}

Despite its unified interface, \texttt{\libname} faces some constraints.
Mirroring industry shifts toward single-framework specialization \cite{debut2025transformers}, our implementation is restricted to PyTorch, precluding direct use by researchers in the JAX or TensorFlow communities.
Furthermore, the high-performance execution of several LTV layers relies on custom CUDA kernels, limiting optimal performance to NVIDIA hardware, and hindering accessibility for alternative backends.

We note that our models match other public implementations on training speed but are slightly slower for inference. 
We attribute this to known CPU overheads in PyTorch inference execution rather than to model-specific design choices.
Though for production workloads, particularly in high batch size and long sequence length regimes, we expect inference performance to be very similar.
Finally, the library lacks native wrappers for established ecosystem tools like Hugging Face \citep{hftransformers}, DeepSpeed \citep{deepspeed}, and FSDP \citep{fsdp}.
Consequently, incorporating these models into large-scale distributed workflows requires the manual development of custom adapter layers.
Beyond ecosystem integrations, there are architectural features we have not yet implemented.
We do not yet provide bidirectional variants of LRNN layers, though the base interface is designed to support them.

Recently there has also been a resurgence in non-linear RNNs like xLSTM~\cite{beck:24xlstm}, while related in capabilities, these methods are orthogonal to our focus and thus have not been implemented.

\bibliography{main}

@article{turing_complete,
  author = {Siegelmann, Hava T. and Sontag, Eduardo D.},
  title = {On the Computational Power of Neural Nets},
  journal = {Journal of Computer and System Sciences},
  year = {1995},
  volume = {50},
  number = {1},
  pages = {132-150}
}

@article{lstm,
  author = {Hochreiter, Sepp and Schmidhuber, J{\"u}rgen},
  title = {Long Short-Term Memory},
  journal = {Neural Computation},
  volume = {9},
  number = {8},
  year = {1997},
  pages = {1735--1780},
  doi = {10.1162/neco.1997.9.8.1735}
}

@inproceedings{Vaswani+2017,
  author = {Vaswani, Ashish and Shazeer, Noam and Parmar, Niki and Uszkoreit, Jakob and Jones, Llion and Gomez, Aidan N. and Kaiser, Łukasz and Polosukhin, Illia},
  title = {Attention is All You Need},
  booktitle = {Advances in Neural Information Processing Systems},
  year = {2017},
  pages = {5998--6008},
  url = {https://arxiv.org/abs/1706.03762}
}

@article{gpt3,
  author = {Brown, Tom B. and Mann, Benjamin and Ryder, Nick and Subbiah, Melanie and Kaplan, Jared and Dhariwal, Prafulla and Neelakantan, Arvind and others},
  title = {Language Models are Few-Shot Learners},
  journal = {Advances in Neural Information Processing Systems},
  year = {2020},
  url = {https://arxiv.org/abs/2005.14165}
}

@inproceedings{gu2022efficiently,
    title={Efficiently Modeling Long Sequences with Structured State Spaces},
    author={Albert Gu and Karan Goel and Christopher Re},
    booktitle={International Conference on Learning Representations},
    year={2022},
    url={https://openreview.net/forum?id=uYLFoz1vlAC}
}

@inproceedings{smith2023simplified,
    title={Simplified State Space Layers for Sequence Modeling},
    author={Jimmy T.H. Smith and Andrew Warrington and Scott Linderman},
    booktitle={The Eleventh International Conference on Learning Representations },
    year={2023},
    url={https://openreview.net/forum?id=Ai8Hw3AXqks}
}

@misc{orvieto2023resurrectingrecurrentneuralnetworks,
      title={Resurrecting Recurrent Neural Networks for Long Sequences}, 
      author={Antonio Orvieto and Samuel L Smith and Albert Gu and Anushan Fernando and Caglar Gulcehre and Razvan Pascanu and Soham De},
      year={2023},
      eprint={2303.06349},
      archivePrefix={arXiv},
      primaryClass={cs.LG},
      url={https://arxiv.org/abs/2303.06349}, 
}

@misc{Schoene2024,
      title={Scalable Event-by-event Processing of Neuromorphic Sensory Signals With Deep State-Space Models}, 
      author={Mark Schöne and Neeraj Mohan Sushma and Jingyue Zhuge and Christian Mayr and Anand Subramoney and David Kappel},
      year={2024},
      eprint={2404.18508},
      archivePrefix={arXiv},
      primaryClass={cs.LG}
}

@inproceedings{gu2024mamba,
    title={Mamba: Linear-Time Sequence Modeling with Selective State Spaces},
    author={Albert Gu and Tri Dao},
    booktitle={First Conference on Language Modeling},
    year={2024},
    url={https://openreview.net/forum?id=tEYskw1VY2}
}

@misc{soydan2024s7selectivesimplifiedstate,
      title={S7: Selective and Simplified State Space Layers for Sequence Modeling}, 
      author={Taylan Soydan and Nikola Zubić and Nico Messikommer and Siddhartha Mishra and Davide Scaramuzza},
      year={2024},
      eprint={2410.03464},
      archivePrefix={arXiv},
      primaryClass={cs.LG},
      url={https://arxiv.org/abs/2410.03464}, 
}

@inproceedings{pei2025let,
    title={Let {SSM}s be ConvNets: State-space Modeling with Optimal Tensor Contractions},
    author={Yan Ru Pei},
    booktitle={The Thirteenth International Conference on Learning Representations},
    year={2025},
    url={https://openreview.net/forum?id=PkpNRmBZ32}
}

@InProceedings{pmlr-v162-goel22a,
  title = 	 {It’s Raw! {A}udio Generation with State-Space Models},
  author =       {Goel, Karan and Gu, Albert and Donahue, Chris and Re, Christopher},
  booktitle = 	 {Proceedings of the 39th International Conference on Machine Learning},
  pages = 	 {7616--7633},
  year = 	 {2022},
  editor = 	 {Chaudhuri, Kamalika and Jegelka, Stefanie and Song, Le and Szepesvari, Csaba and Niu, Gang and Sabato, Sivan},
  volume = 	 {162},
  series = 	 {Proceedings of Machine Learning Research},
  month = 	 {17--23 Jul},
  publisher =    {PMLR},
  url = 	 {https://proceedings.mlr.press/v162/goel22a.html},
}

@inproceedings{pei2025atennuateoptimizedrealtimespeech,
  title     = {{Optimized Real-time Speech Enhancement with Deep SSMs on Raw Audio}},
  author    = {Yan Ru Pei and Ritik Shrivastava and Fnu Sidharth},
  year      = {2025},
  booktitle = {{Interspeech 2025}},
  pages     = {51--55},
  doi       = {10.21437/Interspeech.2025-19},
  issn      = {2958-1796},
}

@misc{ramesh2025lyraefficientexpressivesubquadratic,
      title={Lyra: An Efficient and Expressive Subquadratic Architecture for Modeling Biological Sequences}, 
      author={Krithik Ramesh and Sameed M. Siddiqui and Albert Gu and Michael D. Mitzenmacher and Pardis C. Sabeti},
      year={2025},
      eprint={2503.16351},
      archivePrefix={arXiv},
      primaryClass={cs.LG},
      url={https://arxiv.org/abs/2503.16351}, 
}

@inproceedings{liu2024vmamba,
    title={{VM}amba: Visual State Space Model},
    author={Yue Liu and Yunjie Tian and Yuzhong Zhao and Hongtian Yu and Lingxi Xie and Yaowei Wang and Qixiang Ye and Jianbin Jiao and Yunfan Liu},
    booktitle={The Thirty-eighth Annual Conference on Neural Information Processing Systems},
    year={2024},
    url={https://openreview.net/forum?id=ZgtLQQR1K7}
}

@misc{schöne2024streamuniversalstatespacemodel,
      title={STREAM: A Universal State-Space Model for Sparse Geometric Data}, 
      author={Mark Schöne and Karan Bania and Yash Bhisikar and Khaleelulla Khan Nazeer and Christian Mayr and Anand Subramoney and David Kappel},
      year={2024},
      eprint={2411.12603},
      archivePrefix={arXiv},
      primaryClass={cs.CV},
      url={https://arxiv.org/abs/2411.12603}, 
}

@article{tay2021long,
  author = {Tay, Yi and Dehghani, Mostafa and Bahri, Dara and Metzler, Donald},
  title = {Long Range Arena: A Benchmark for Efficient Transformers},
  journal = {arXiv preprint arXiv:2011.04006},
  year = {2021},
  url = {https://arxiv.org/abs/2011.04006}
}

@inproceedings{mamba3d,
    title={Mamba3D: Enhancing Local Features for 3D Point Cloud Analysis via State Space Model},
    author={Xu Han and Yuan Tang and Zhaoxuan Wang and Xianzhi Li},
    booktitle={ACM Multimedia 2024},
    year={2024},
    url={https://openreview.net/forum?id=Tl13I7b3Ao}
}

@article{beltagy2020longformer,
  author = {Beltagy, Iz and Peters, Matthew E. and Cohan, Arman},
  title = {Longformer: The Long-Document Transformer},
  journal = {arXiv preprint arXiv:2004.05150},
  year = {2020},
  url = {https://arxiv.org/abs/2004.05150}
}

@article{wang2020linformer,
  author = {Wang, Sinong and Li, Belinda Z. and Khabsa, Madian and Fang, Han and Ma, Hao},
  title = {Linformer: Self-Attention with Linear Complexity},
  journal = {arXiv preprint arXiv:2006.04768},
  year = {2020},
  url = {https://arxiv.org/abs/2006.04768}
}

@article{kitaev2020reformer,
  author = {Kitaev, Nikita and Kaiser, Łukasz and Levskaya, Anselm},
  title = {Reformer: The Efficient Transformer},
  journal = {arXiv preprint arXiv:2001.04451},
  year = {2020},
  url = {https://arxiv.org/abs/2001.04451}
}

@article{dao2022flashattention,
  author = {Dao, Tri and Fu, Daniel Y. and Ermon, Stefano and Rudra, Atri and Ré, Christopher},
  title = {FlashAttention: Fast and Memory-Efficient Exact Attention with IO-Awareness},
  journal = {arXiv preprint arXiv:2205.14135},
  year = {2022},
  url = {https://arxiv.org/abs/2205.14135}
}

@article{kusupati2022matryoshka,
  author = {Kusupati, Aditya and Bhatt, Gantavya and Rege, Aniket and Wallingford, Matthew and Sinha, Aditya and Ramanujan, Vivek and Howard-Snyder, William and Chen, Kaifeng and Kakade, Sham and Jain, Prateek and Farhadi, Ali},
  title = {Matryoshka Representation Learning},
  journal = {arXiv preprint arXiv:2205.13147},
  year = {2022},
  url = {https://arxiv.org/abs/2205.13147}
}

@article{leviathan2023fast,
  author = {Leviathan, Yaniv and Kalman, Matan and Matias, Yossi},
  title = {Fast Inference from Transformers via Speculative Decoding},
  journal = {arXiv preprint arXiv:2211.17192},
  year = {2023},
  url = {https://arxiv.org/abs/2211.17192}
}

@misc{de2024griffinmixinggatedlinear,
      title={Griffin: Mixing Gated Linear Recurrences with Local Attention for Efficient Language Models}, 
      author={Soham De and Samuel L. Smith and Anushan Fernando and Aleksandar Botev and George Cristian-Muraru and Albert Gu and Ruba Haroun and Leonard Berrada and Yutian Chen and Srivatsan Srinivasan and Guillaume Desjardins and Arnaud Doucet and David Budden and Yee Whye Teh and Razvan Pascanu and Nando De Freitas and Caglar Gulcehre},
      year={2024},
      eprint={2402.19427},
      archivePrefix={arXiv},
      primaryClass={cs.LG},
      url={https://arxiv.org/abs/2402.19427}, 
}

@article{kwon2023efficient,
  author = {Kwon, Woosuk and Li, Zhuohan and Zhuang, Siyuan and Sheng, Ying and Zheng, Lianmin and Yu, Cody Hao and Gonzalez, Joseph E. and Zhang, Hao and Stoica, Ion},
  title = {Efficient Memory Management for Large Language Model Serving with PagedAttention},
  journal = {arXiv preprint arXiv:2309.06180},
  year = {2023},
  url = {https://arxiv.org/abs/2309.06180}
}

@inproceedings{hftransformers,
    title = "Transformers: State-of-the-Art Natural Language Processing",
    author = "Wolf, Thomas  and
      Debut, Lysandre  and
      Sanh, Victor  and
      Chaumond, Julien  and
      Delangue, Clement  and
      Moi, Anthony  and
      Cistac, Pierric  and
      Rault, Tim  and
      Louf, Remi  and
      Funtowicz, Morgan  and
      Davison, Joe  and
      Shleifer, Sam  and
      von Platen, Patrick  and
      Ma, Clara  and
      Jernite, Yacine  and
      Plu, Julien  and
      Xu, Canwen  and
      Le Scao, Teven  and
      Gugger, Sylvain  and
      Drame, Mariama  and
      Lhoest, Quentin  and
      Rush, Alexander",
    editor = "Liu, Qun  and
      Schlangen, David",
    booktitle = "Proceedings of the 2020 Conference on Empirical Methods in Natural Language Processing: System Demonstrations",
    month = oct,
    year = "2020",
    address = "Online",
    publisher = "Association for Computational Linguistics",
    url = "https://aclanthology.org/2020.emnlp-demos.6/",
    doi = "10.18653/v1/2020.emnlp-demos.6",
    pages = "38--45"
}

@online{debut2025transformers,
  author    = {Lysandre Debut},
  title     = {LinkedIn Post},
  year      = {2025},
  month     = {June},
  url       = {https://www.linkedin.com/feed/update/urn:li:activity:7338966863403528192/},
  urldate   = {2025-10-25}
}

@misc{zucchet2023minimalLRU,
  author = {Nicolas Zucchet},
  title = {minimal-LRU: JAX implementation of the Linear Recurrent Unit},
  year = {2023},
  howpublished = {\url{https://github.com/NicolasZucchet/minimal-LRU}},
  note = {GitHub repository}
}

@article{fsdp,
author = {Zhao, Yanli and Gu, Andrew and Varma, Rohan and Luo, Liang and Huang, Chien-Chin and Xu, Min and Wright, Less and Shojanazeri, Hamid and Ott, Myle and Shleifer, Sam and Desmaison, Alban and Balioglu, Can and Damania, Pritam and Nguyen, Bernard and Chauhan, Geeta and Hao, Yuchen and Mathews, Ajit and Li, Shen},
title = {PyTorch FSDP: Experiences on Scaling Fully Sharded Data Parallel},
year = {2023},
issue_date = {August 2023},
publisher = {VLDB Endowment},
volume = {16},
number = {12},
issn = {2150-8097},
url = {https://doi.org/10.14778/3611540.3611569},
doi = {10.14778/3611540.3611569},
journal = {Proc. VLDB Endow.},
month = aug,
pages = {3848–3860},
numpages = {13}
}

@inproceedings{deepspeed,
author = {Rasley, Jeff and Rajbhandari, Samyam and Ruwase, Olatunji and He, Yuxiong},
title = {DeepSpeed: System Optimizations Enable Training Deep Learning Models with Over 100 Billion Parameters},
year = {2020},
isbn = {9781450379984},
publisher = {Association for Computing Machinery},
address = {New York, NY, USA},
url = {https://doi.org/10.1145/3394486.3406703},
doi = {10.1145/3394486.3406703},
booktitle = {Proceedings of the 26th ACM SIGKDD International Conference on Knowledge Discovery \& Data Mining},
pages = {3505–3506},
numpages = {2},
location = {Virtual Event, CA, USA},
series = {KDD '20}
}

@inproceedings{beck:24xlstm,
  title = {xLSTM: Extended Long Short-Term Memory}, 
  author = {Maximilian Beck and Korbinian Pöppel and Markus Spanring and Andreas Auer and Oleksandra Prudnikova and Michael Kopp and Günter Klambauer and Johannes Brandstetter and Sepp Hochreiter},
  booktitle = {Thirty-eighth Conference on Neural Information Processing Systems},
  year = {2024},
  url = {https://arxiv.org/abs/2405.04517}, 
}
\onecolumn

\newpage


\begin{appendices}

\section{Benchmarks}
\label{app:benchmarks}

\begin{figure}[!h]
    \centering
    \includegraphics[width=\linewidth]{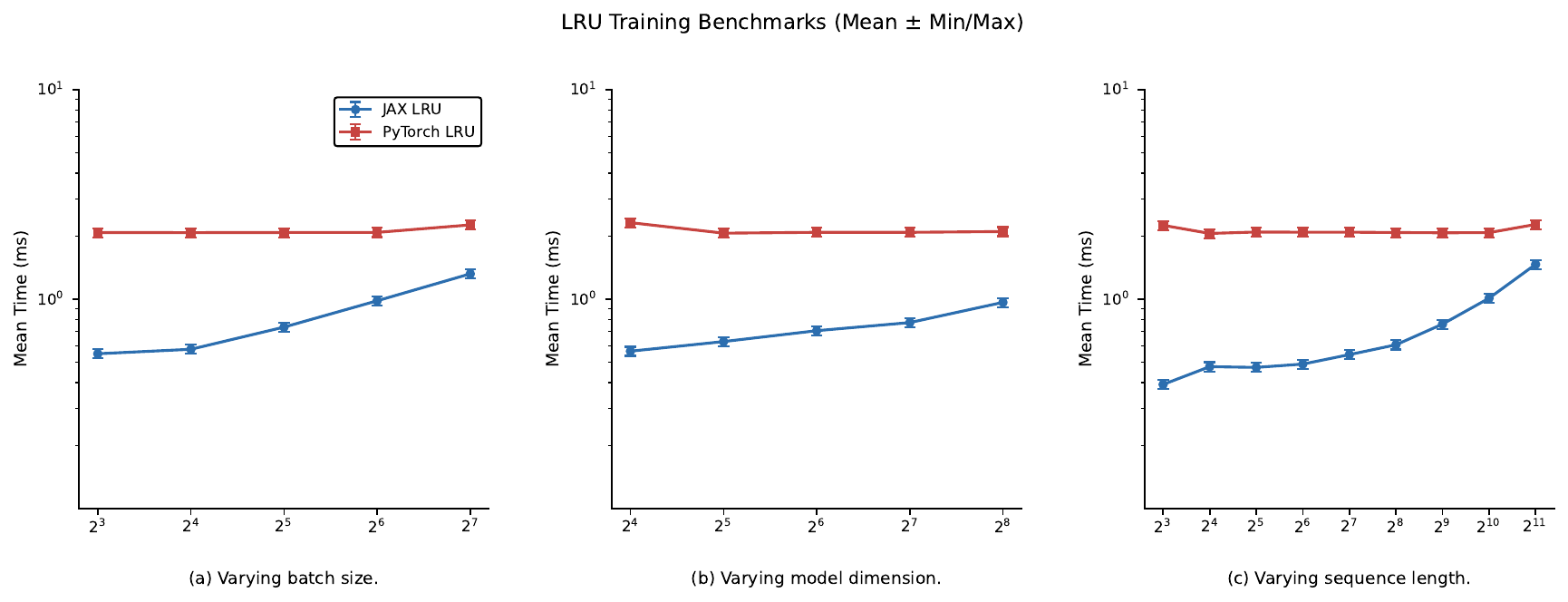}
    \caption{LRU Training Benchmarks.}
    \label{fig:placeholder}
\end{figure}
\begin{figure}[!h]
    \centering
    \includegraphics[width=\linewidth]{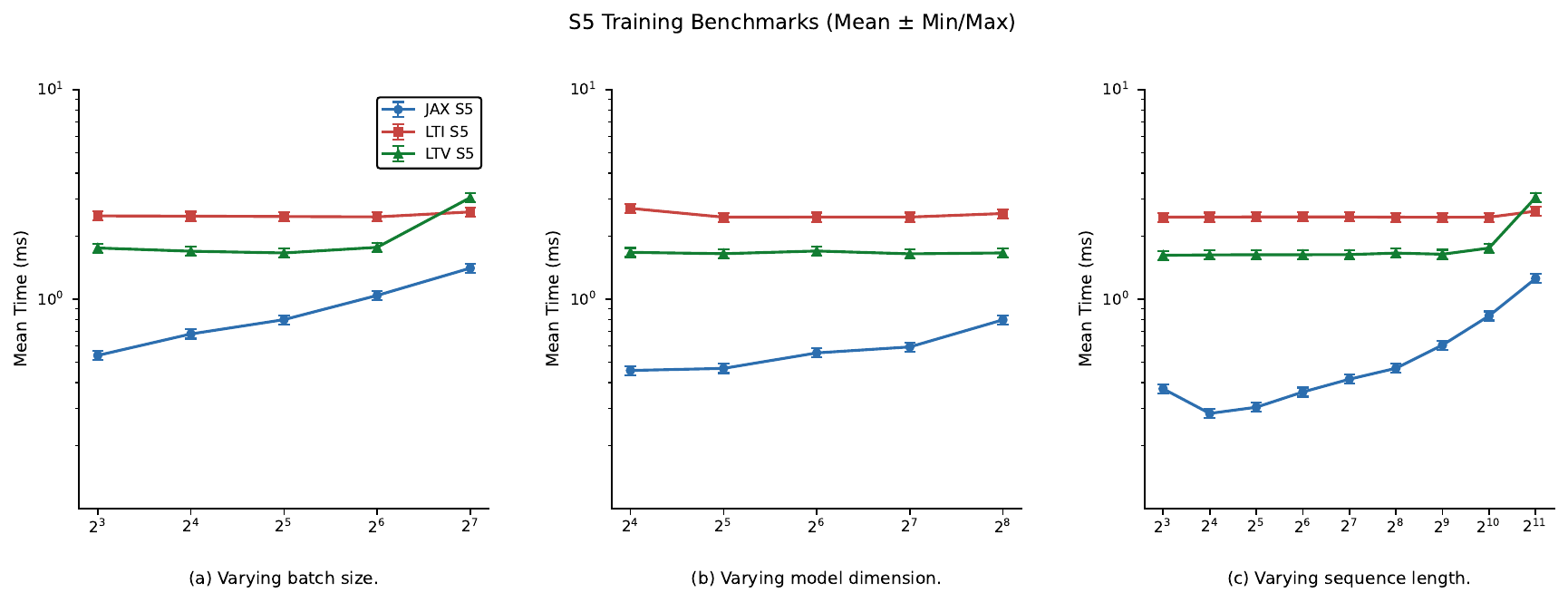}
    \caption{S5 Training Benchmarks.}
    \label{fig:placeholder}
\end{figure}
\begin{figure}[!h]
    \centering
    \includegraphics[width=\linewidth]{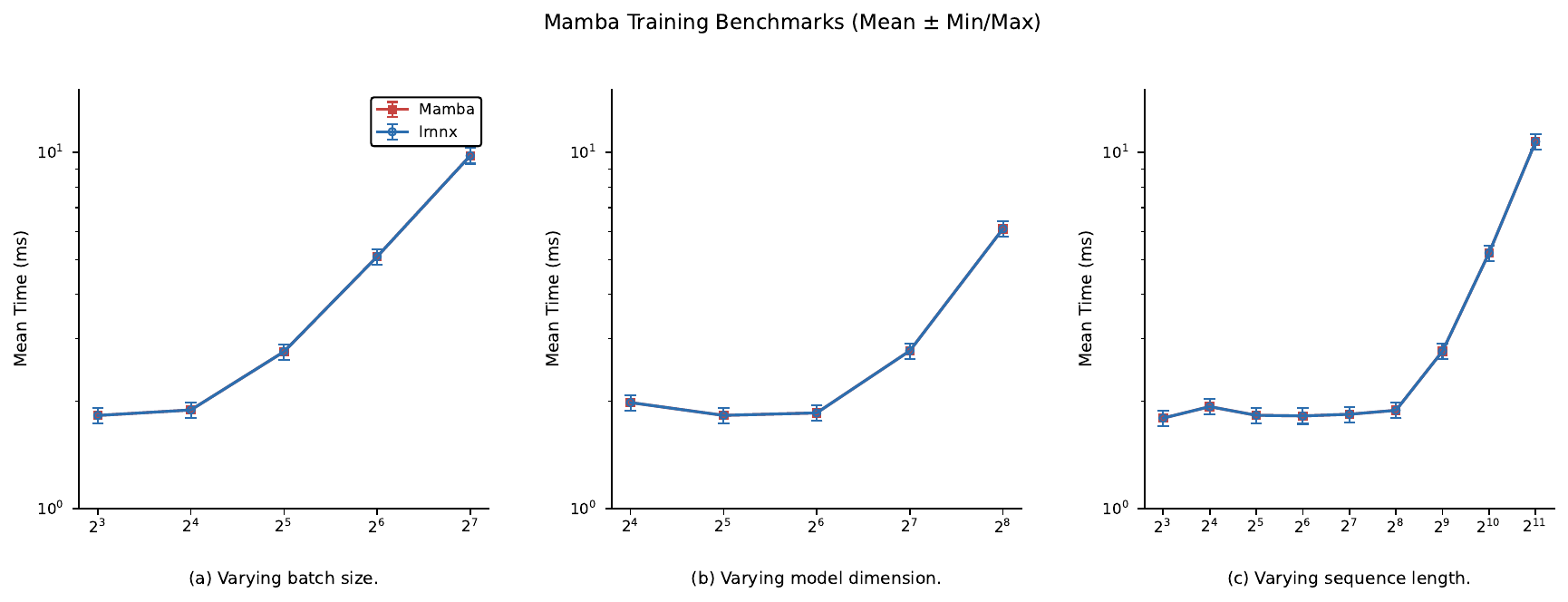}
    \caption{Mamba Training Benchmarks.}
    \label{fig:placeholder}
\end{figure}
\begin{figure}[!h]
    \centering
    \includegraphics[width=\linewidth]{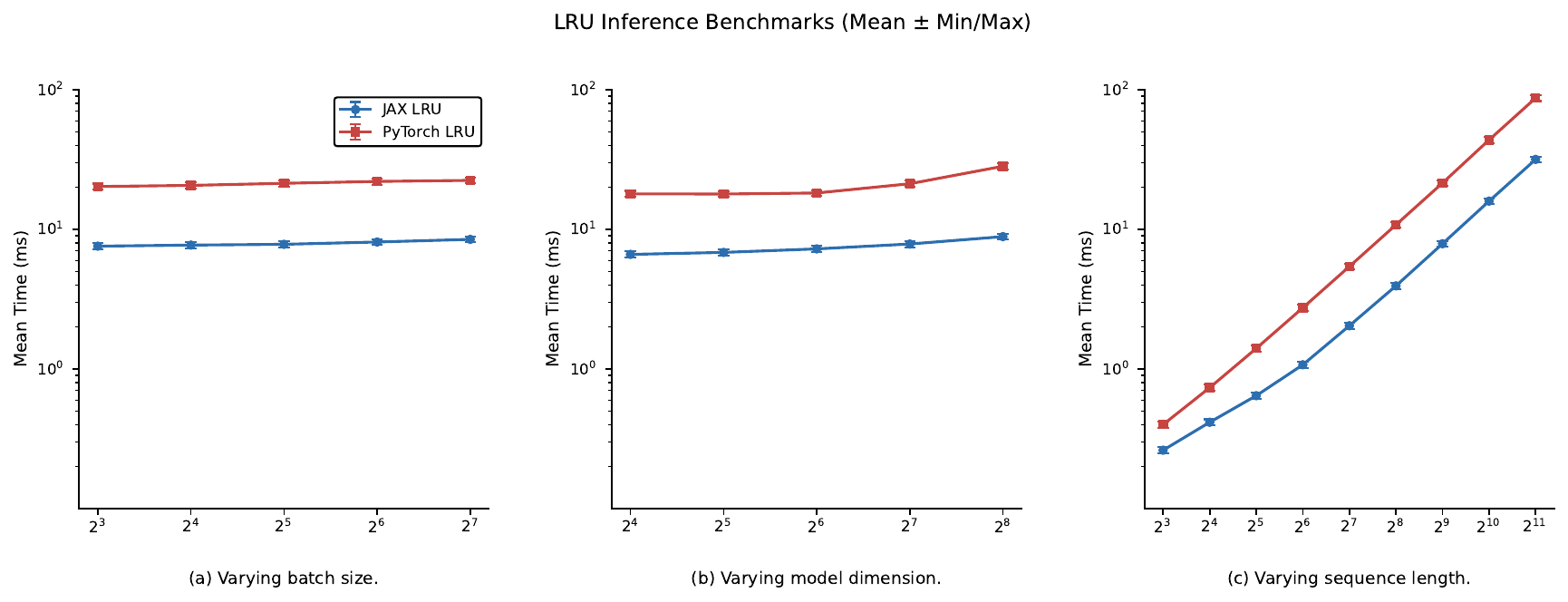}
    \caption{LRU Inference Benchmarks.}
    \label{fig:placeholder}
\end{figure}
\begin{figure}[!h]
    \centering
    \includegraphics[width=\linewidth]{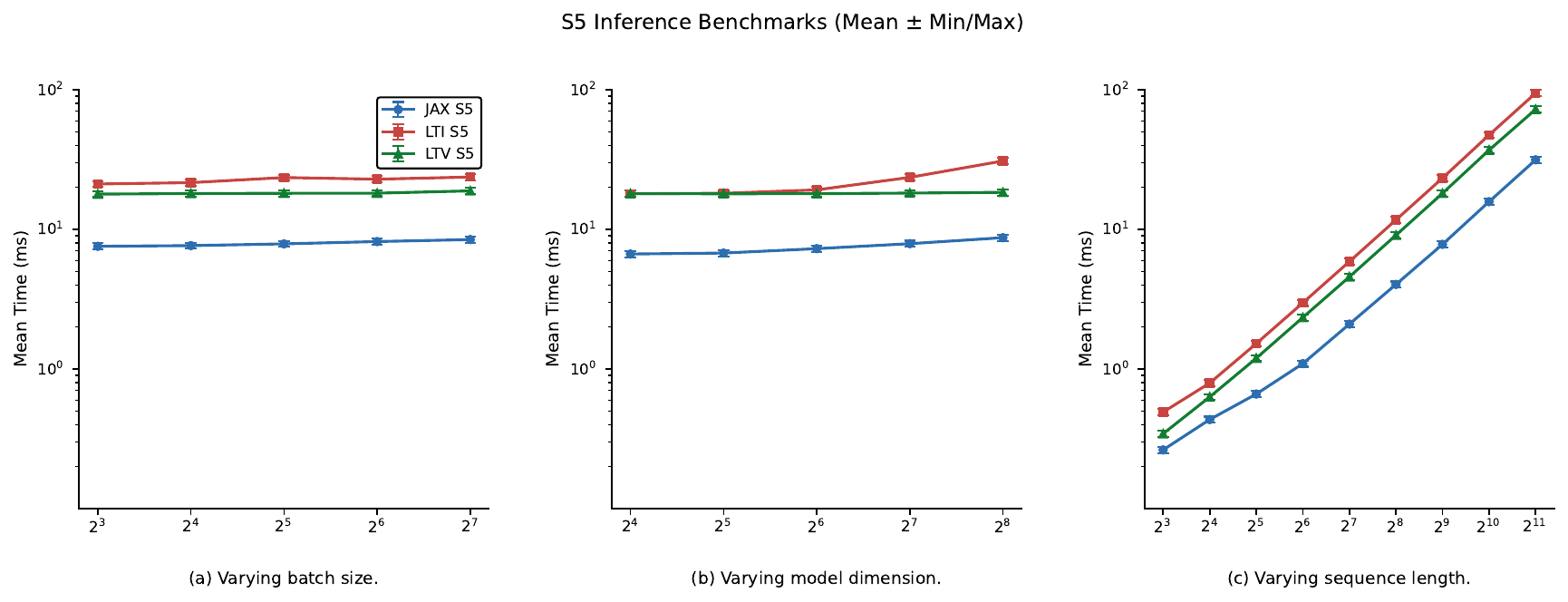}
    \caption{S5 Inference Benchmarks.}
    \label{fig:placeholder}
\end{figure}
\begin{figure}[!h]
    \centering
    \includegraphics[width=\linewidth]{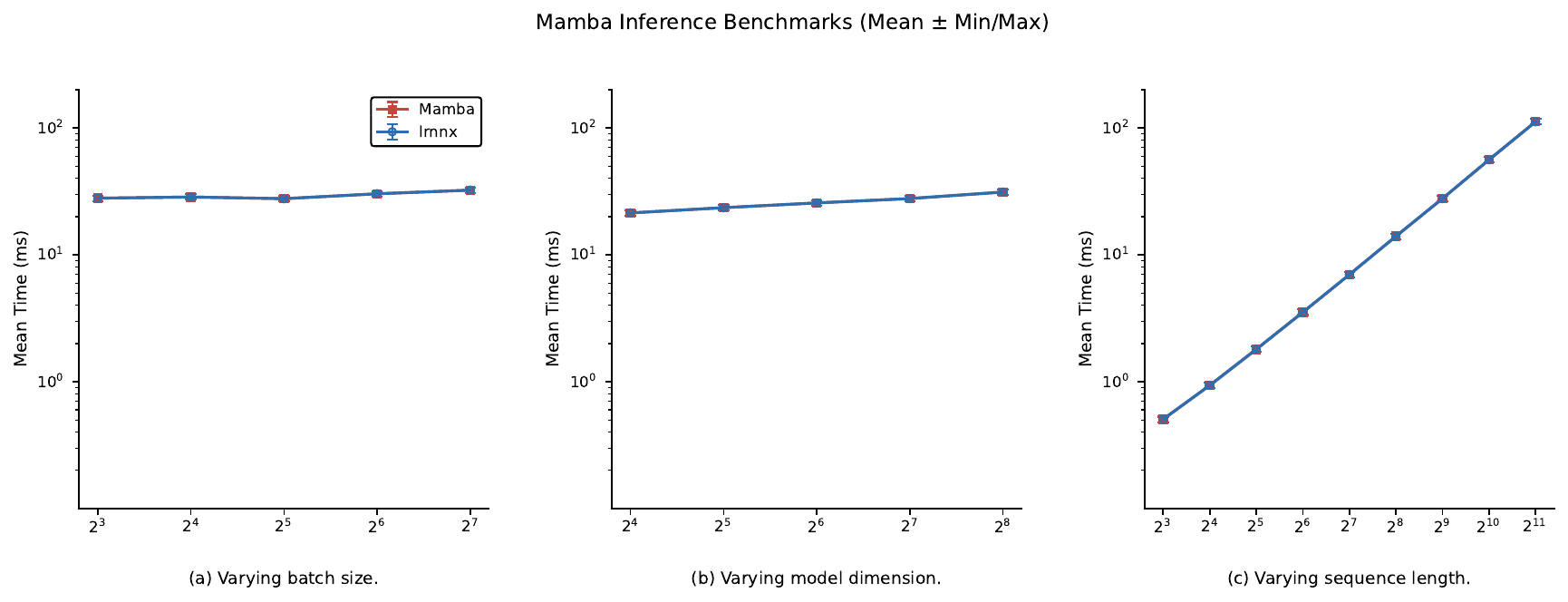}
    \caption{Mamba Inference Benchmarks.}
    \label{fig:placeholder}
\end{figure}

\end{appendices}

\end{document}